\documentclass[sigconf]{acmart}

\def\BibTeX{{\rm B\kern-.05em{\sc i\kern-.025em b}\kern-.08emT\kern-.1667em\lower.7ex\hbox{E}\kern-.125emX}}
    
\copyrightyear{2018}
\acmYear{2018}
\setcopyright{acmlicensed}
\acmPrice{15.00}
\acmDOI{10.1145/1122445.1122456}
\acmISBN{978-1-4503-9999-9/18/06}

\begin{document}

%
\title{Emotionally-Aware Chatbots: A Survey}

%
\author{Endang Wahyu Pamungkas}
\email{pamungka@di.unito.it}
\affiliation{%
  \institution{University of Turin}
  \streetaddress{via Pessinetto, 12}
  \city{Turin}
  \state{Italy}
}

%
\renewcommand{\shortauthors}{Pamungkas.}

%
\begin{abstract}
Textual conversational agent or chatbots' development gather tremendous traction from both academia and industries in recent years. Nowadays, chatbots are widely used as an agent to communicate with a human in some services such as booking assistant, customer service, and also a personal partner. The biggest challenge in building chatbot is to build a humanizing machine to improve user engagement. Some studies show that emotion is an important aspect to humanize machine, including chatbot. In this paper, we will provide a systematic review of approaches in building an emotionally-aware chatbot (EAC). As far as our knowledge, there is still no work focusing on this area. We propose three research question regarding EAC studies. We start with the history and evolution of EAC, then several approaches to build EAC by previous studies, and some available resources in building EAC. Based on our investigation, we found that in the early development, EAC exploits a simple rule-based approach while now most of EAC use neural-based approach. We also notice that most of EAC contain emotion classifier in their architecture, which utilize several available affective resources. We also predict that the development of EAC will continue to gain more and more attention from scholars, noted by some recent studies propose new datasets for building EAC in various languages.
\end{abstract}

%
%


%

%
\maketitle

\section{Introduction}

Conversational agents or dialogue systems development are gaining more attention from both industry and academia \cite{hancock2019learning,fang2018sounding} in the latest years. Some works tried to model them into domain-specific tasks such as customer service \cite{hu2018touch,cui2017superagent}, and shopping assistance \cite{kothari2017chatbots}. Other works design a multi-purpose agents such as SIRI\footnote{\url{https://www.apple.com/es/siri/}}, Amazon Alexa\footnote{\url{https://developer.amazon.com/alexa}}, and Google Assistance\footnote{\url{https://assistant.google.com/}}. This domain is a well-researched area in Human-Computer Interaction research community but still become a hot topic now. The main development focus right now is to have an intelligent and humanizing machine to have a better engagement when communicating with human \cite{go2019humanizing}. Having a better engagement will lead to higher user satisfaction, which becomes the main objective from the industry perspective.

In this study, we will only focus on textual conversational agent or chatbot, a conversational artificial intelligence which can conduct a textual communication with a human by exploiting several natural language processing techniques. There are several approaches used to build a chatbot, start by using a simple rule-based approach \cite{ritter2011data,al2011ontbot} until more sophisticated one by using neural-based technique \cite{qiu2017alime,serban2017deep}. Nowadays, chatbots are mostly used as customer service such as booking systems \cite{wen2017network,peng2017composite}, shopping assistance \cite{cui2017superagent} or just as conversational partner such as Endurance \footnote{\url{https://bit.ly/2RGM40S}} and Insomnobot \footnote{\url{http://insomnobot3000.com/}}. Therefore, there is a significant urgency to humanize chatbot for having a better user-engagement. Some works were already proposed several approaches to improve chatbot's user-engagement, such as building a context-aware chatbot \cite{sordoni-etal-2015-neural} and injecting personality into the machine \cite{zhang2018personalizing}. Other works also try to incorporate affective computing to build emotionally-aware chatbots \cite{hu2018touch,rashkin2018know,zhou2018emotional}.

Some existing studies shows that adding emotion information into dialogue systems is able to improve user-satisfaction \cite{yu2015ticktock, prendinger2005empathic}. Emotion information contribute to a more positive interaction between machine and human, which lead to reduce miscommunication \cite{martinovski2003breakdown}. Some previous studies also found that using affect information can help chatbot to understand users' emotional state, in order to generate better response \cite{polzin2000emotion}. Not only emotion, another study also introduce the use of tones to improve satisfactory service. For instance, using empathetic tone is able to reduces user stress and results in more engagement. \cite{hu2018touch} found that tones is an important aspect in building customer care chatbot. They discover eight different tones including anxious, frustrated, impolite, passionate, polite, sad, satisfied, and empathetic.

In this paper, we will try to summarize some previous studies which focus on injecting emotion information into chatbots, on discovering recent issues and barriers in building engaging emotionally-aware chatbots. Therefore, we propose some research questions to have a better problem definition:
\begin{enumerate}
  \item[RQ1] How to incorporate emotion information in building an emotionally-aware chatbot?
  \item[RQ2] What are available resources that can be used in building emotionally-aware chatbots?
  \item[RQ3] How to evaluate the performance of emotionally-aware chatbots?
\end{enumerate}
This paper will be organized as follows: Section 2 introduces the history of the relation between affective information with chatbots. Section 3 outline some works which try to inject affective information into chatbots. Section 4 summarizes some affective resources which can be utilized to provide affective information. Then, Section 5 describes some evaluation metric that already applied in some previous works related to emotionally-aware chatbots. Last Section 6 will conclude the rest of the paper and provide a prediction of future development in this research direction based on our analysis.

\section{History of Emotionally-Aware Chatbot}

The early development of chatbot was inspired by Turing test in 1950 \cite{turing2009computing}. Eliza was the first publicly known chatbot, built by using simple hand-crafted script \cite{weizenbaum1966eliza}. Parry \cite{colby2013artificial} was another chatbot which successfully passed the Turing test. Similar to Eliza, Parry still uses a rule-based approach but with a better understanding, including the mental model that can stimulate emotion. Therefore, Parry is the first chatbot which involving emotion in its development. Also, worth to be mentioned is ALICE (Artificial Linguistic Internet Computer Entity), a customizable chatbot by using Artificial Intelligence Markup Language (AIML). Therefore, ALICE still also use a rule-based approach by executing a pattern-matcher recursively to obtain the response. Then in May 2014, Microsoft introduced XiaoIce \cite{zhou2018design}, an empathetic social chatbot which is able to recognize users' emotional needs. XiaoIce can provide an engaging interpersonal communication by giving encouragement or other affective messages, so that can hold human attention during communication.

Nowadays, most of chatbots technologies were built by using neural-based approach. Emotional Chatting Machine (ECM) \cite{zhou2018emotional} was the first works which exploiting deep learning approach in building a large-scale emotionally-aware conversational bot. Then several studies were proposed to deal with this research area by introducing emotion embedding representation \cite{shantala2018neural,asghar2018affective,colombo2019affect} or modeling as reinforcement learning problem \cite{sun2018emotional,li2019reinforcement}. Most of these studies used encoder-decoder architecture, specifically sequence to sequence (seq2seq) learning. Some works also tried to introduce a new dataset in order to have a better gold standard and improve system performance. \cite{rashkin2018know} introduce EMPATHETICDIALOGUES dataset, a novel dataset containing 25k conversations include emotional contexts information to facilitate training and evaluating the textual conversational system. Then, work from \cite{hu2018touch} produce a dataset containing 1.5 million Twitter conversation, gathered by using Twitter API from customer care account of 62 brands across several industries. This dataset was used to build tone-aware customer care chatbot. Finally, \cite{lubis2019positive} tried to enhance SEMAINE corpus \cite{mckeown2012semaine} by using crowdsourcing scenario to obtain a human judgement for deciding which response that elicits positive emotion. Their dataset was used to develop a chatbot which captures human emotional states and elicits positive emotion during the conversation.

\section{Building Emotionally-Aware Chatbot (EAC)}
As we mentioned before that emotion is an essential aspect of building humanize chatbot. The rise of the emotionally-aware chatbot is started by Parry \cite{colby2013artificial} in early 1975. Now, most of EAC development exploits neural-based model. In this section, we will try to review previous works which focus on EAC development. Table~\ref{system-review} summarizes this information includes the objective and exploited approach of each work. In early development, EAC is designed by using a rule-based approach. However, in recent years mostly EAC exploit neural-based approach. Studies in EAC development become a hot topic start from 2017, noted by the first shared task in Emotion Generation Challenge on NLPCC 2017 \cite{huang2017overview}. Based on Table~\ref{system-review} this research line continues to gain massive attention from scholars in the latest years.

Based on Table~\ref{system-review}, we can see that most of all recent EAC was built by using encoder-decoder architecture with sequence-to-sequence learning. These seq2seq learning models maximize the likelihood of response and are prepared to incorporate rich data to generate an appropriate answer. Basic seq2seq architecture structured of two recurrent neural networks (RNNs), one as an encoder processing the input and one as a decoder generating the response. long short term memory (LSTM) or gated recurrent unit (GRU) was the most dominant variant of RNNs which used to learn the conversational dataset in these models. Some studies also tried to model this task as a reinforcement learning task, in order to get more generic responses and let the chatbot able to achieve successful long-term conversation. Attention mechanism was also introduced in this report\footnote{\url{http://web.stanford.edu/class/cs224s/reports/Honghao_Wei.pdf}}. This mechanism will allow the decoder to focus only on some important parts in the input at every decoding step.

Another vital part of building EAC is emotion classifier to detect emotion contained in the text to produce a more meaningful response. Emotion detection is a well-established task in natural language processing research area. This task was promoted in two latest series of SemEval-2018 (Task 1) and SemEval-2019 (Task 3). Some tasks were focusing on classifying utterance into several categories of emotion \cite{mohammad2018semeval}. However, there is also a task which trying to predict the emotion intensities contained in the text \cite{mohammad2017wassa}. In the early development of emotion classifier, most of the studies proposed to use traditional machine-learning approach. However, the neural-based approach is able to gain better performance, which leads more scholars to exploit it to deal with this task. In chatbot, the system will generate several responses based on several emotion categories. Then the system will respond with the most appropriate emotion based on emotion detected on posted utterance by emotion classifier. Based on Table~\ref{system-review}, studies have different emotion categories based on their focus and objective in building chatbots.

\begin{table*}
\begin{center}
\begin{tabular}{ p{3.5cm}p{1cm}p{4.5cm}p{7cm}}
 \hline
\textbf{Authors} & \textbf{Year} & \textbf{Focus} & \textbf{Approach}\\
\hline
Kenneth Mark Colby \cite{colby2013artificial} & 1975 & Designing chatbot behave like paranoid person & Rule-based approach with capability to simulate emotion.\\
Zhou et. al. \cite{zhou2018design} & 2014 & Building empathetic chatbot for social interaction & Seq2seq learning with GRU-RNN model which takes into account intelligent quotient (IQ) and emotion quotien (EQ).\\
Zhou et. al. \cite{zhou2018emotional} & 2018 & Building emotion chatting machine (ECM) for large scale conversation generation & Seq2seq learning with GRU which incorporates emotion detection to capture implicit internal emotion states.\\
Colombo et. al. \cite{colombo2019affect} & 2019 & Developing affect-driven dialogue system which generates emotional responses in a controlled manner & Seq2seq with GRU which incorporates emotion classifier, also affective re-ranking in the last step to produce the response.\\
Li et. al. \cite{li2019reinforcement} & 2019 & Building conversational system  emotional by using editing constraints to generate more meaningful and customizable emotional replies & Encoder-decoder architecture with reinforcement learning approach by using asynchronous decoder which uses keyword predictor to predict the topic and emotion editor to produce emotional-embedded response.\\
Zhang et. al. \cite{zhang2017building} & 2017 & Building emotional conversation systems which produces several responses for every emotion category & Multi-task seq2seq model with GRU which utilize bidirectional long short term memory (Bi-LSTMs) as emotion classifier.\\
Lubis et. al. \cite{lubis2019positive} & 2019 & Building a fully data driven chat-oriented dialogue system that can dynamically mimic affective human interactions & Proposing a seq2seq response generator, includes emotion encoder which is trained jointly with the entire network to encode and maintain the emotional context throughout the dialogue, but focusing only on positive emotion\\
Ashgar et. al. \cite{asghar2018affective} & 2018 & Building open-domain neural dialogue models by augmenting them with affective intelligence. & Encoder-decoder architecture with LSTM model using cognitively engineered affective word embedding, and also affectively diverse beam search for decoding\\
Sun et. al. \cite{sun2018emotional} & 2018 & Developing conversation generation which addresses the emotional factor by changing the model's input. & Encoder-decoder framework based on LSTM using three inputs, a
sequence without an emotional category, a sequence with an emotional category for the input sentence, and a sequence with an emotional category for the output responses.\\
Hu et. al. \cite{hu2018touch} & 2018 & Building a tone-aware chatbot for customer care on social media. & Encoder-decoder architecture with LSTM, which the decoder is modified to handle meta information by add an embedding vector as tone indicator to produce several tone responses including passionate, empathetic, and neutral. \\
Sun et. al. \cite{sun2018emotional} & 2018 & Building emotional human machine conversation chatbot that be able to understands the user's emotions, and consider the user's emotions then give a satisfactory response. & Model this problem as reinforcement learning task by proposing a new neural model based on Seq2Seq model which uses generative adversarial network (GAN) called EM-SeqGAN. \\
Shantala et. al. \cite{shantala2018neural} & 2018 & Building neural dialogue system which addresses emotional aspect. & Seq2Seq model with LSTM by using a novel emotion embedding. \\
Zhong et. al. \cite{zhong2018affect} & 2018 & Building affect-rich open domain human-machine conversation system. & Extends Seq2Seq model and adopts VAD (Valence, Arousal and Dominance) affective notations. Proposed model also considers the effect of negators and intensifiers via a novel affective attention mechanism.\\
Catania et. al. \cite{catania2019cork} & 2019 & Building a modular framework to facilitate and accelerate the realization and the maintenance of intelligent Conversational Agents with both rational and emotional capabilities. & The model consists of several modules including intent predictor, sentiment analysis, emotion analysis, topic analysis, and profiling module.\\
\hline
\end{tabular}
\caption{\label{system-review} Summarization of the Proposed Approaches for Emotionally-Aware Chatbot Development.}
\end{center}
\end{table*}

\section{Resource for Building EAC}
In this section, we try to investigate the available resources in building EAC. As other artificial intelligent agents, building chatbot also needs a dataset to be learned, to be able to produce a meaningful conversation as a human-like agent. Therefore, some studies propose dataset which contains textual conversation annotated by different emotion categories. Table~\ref{dataset-review} summarizes the available dataset that found in recent years. We categorize the dataset based on the language, source of the data, and further description which contains some information such as annotation approach, size of instances, and emotion labels. All datasets were proposed during 2017 and 2018, started by dataset provided by NLPCC 2017 Shared Task on Emotion Generation Challenge organizers. This dataset gathered from Sina Weibo social media \footnote{\url{https://www.weibo.com/login.php}}, so it consists of social conversation in Chinese. Based on our study, all of the datasets that we discover are only available in two languages, English and Chinese. However, the source of these datasets is very diverse such as social media (Twitter, Sina Weibo, and Facebook Message), online content, and human writing through crowdsourcing scenario. Our investigation found that every dataset use different set of emotion label depends on its focus and objective in building the chatbot.

As we mentioned in the previous section, that emotion classifier is also an integral part of the emotionally-aware chatbot. In building emotion classifier, there are also available several affective resources, which already widely used in the emotion classification task. Table~\ref{affective-resources-review} shows the available affective resources we discovered, range from old resources such as LIWC, ANEW, and DAL, until the modern version such as DepecheMood and EmoWordNet. Based on some prior studies, emotion can be classified into two fundamental viewpoints, including discrete categories and dimensional models. Meanwhile, this feature view emotion as a discrete category which differentiates emotion into several primary emotion. The most popular was proposed by \cite{ekman1992argument} that differentiate emotion into six categories including anger, disgust, fear, happiness, sadness, and surprise. Three different resources will be used to get emotion contained in the tweet, including EmoLex, EmoSenticNet, LIWC, DepecheMood, and EmoWordNet. Dimensional model is another viewpoint of emotion that define emotion according to one or more dimension. Based on the dimensional approach, emotion is a coincidence value on some different strategic dimensions. We will use two various lexical resources to get the information about the emotional dimension of each tweet, including Dictionary of Affect in Language (DAL), ANEW, and NRC VAD. 

\begin{table*}
\begin{center}
\begin{tabular}{ p{3cm}p{1cm}p{1.5cm}p{1.5cm}p{9cm}}
 \hline
\textbf{Authors} & \textbf{Year} & \textbf{Language} & \textbf{Source} & \textbf{Description}\\
\hline
Zhou et. al. \cite{zhou2018emotional} & 2018 & Chinese & Weibo & This dataset used STC dataset \cite{shang2015neural}, and performed automatic annotation by using the best performing classifier trained on NLPCC 2013 \footnote{\url{http://tcci.ccf.org.cn/conference/2013/}} and NLPCC 2014 \footnote{\url{http://tcci.ccf.org.cn/conference/2014/}} datasets. This dataset contains 217,905 conversations where each response annotated by 6 emotion labels. \\
Rashkin et. al. \cite{rashkin2018know} & 2018 & English & ParlAI Platform & This dataset is built by using crowd-sourcing scenario using ParlAI which involves 810 different participants. This dataset contains 24,850 conversations/prompts where each conversation annotated by 32 emotion labels. \\
Huang et. al. \cite{huang2017overview} & 2017 & Chinese & Weibo & Similar to \cite{zhou2018emotional}, the training data is automatically annotated by using trained emotion classifier. This dataset contains more than 1 million conversations as training and 200 conversation which was manually annotated as test set, where each response labeled by 5 emotion labels. \\
Hu et. al. \cite{hu2018touch} & 2018 & English & Twitter & This dataset gathered from Twitter based on customer care conversation in 62 brands across different industries. 500 conversation was chosen randomly and annotated by 8 major tones using CrowdFlower \footnote{\url{https://www.figure-eight.com/}} platform\\
Lubis et. al. \cite{lubis2018eliciting} & 2018 & English & Crowd source & This dataset was built by enhancing SEMAINE dataset \cite{mckeown2012semaine}, adding a label on instance which elicit positive emotions (8 labels including alert, excited, elated, happy, content, serene, relaxed, calm). The final dataset contains 2,349 conversations.\\
Li et. al. \cite{li2017dailydialog} & 2017 & English & Online Website & This dataset (called ``DailyDialog'') is crawled from various websites which serve for English learner to practice English dialog in daily life. DailyDialog contains 13,118 dialogs, and was manually annotated with communication intention and emotion information. They use six primary basic emotion (Anger, Disgust, Fear, Happiness, Sadness, Surprise) to label the conversations.\\
Huang et. al. \cite{huang2018emotionpush} & 2018 & English & Facebook Message & This dataset was collected from private conversation on Facebook messenger application. 8,818 messages were manually annotated by using seven emotion labels including joy, anger, sadness, anticipation, tired, fear and neutral.\\
\hline
\end{tabular}
\caption{\label{dataset-review} Summarization of dataset available for emotionally-aware chatbot.}
\end{center}
\end{table*}

\begin{table*}
\begin{center}
\begin{tabular}{ p{3.5cm}p{3.5cm}p{9cm}}
 \hline
\textbf{Authors} & \textbf{Name}  & \textbf{Descroption}\\
\hline
Pennebaker et. al. \cite{pennebaker2001linguistic} & Linguistic Inquiry and Word Count (LIWC) & LIWC is a computer application that offers an efficient instrument for word-by-word study of the emotional, cognitive, and structural elements in English. LIWC also provide a dictionary which has 4500 words distributed into 64 different emotional categories including positive and negative.\\
Mohammad et. al. \cite{mohammad2013crowdsourcing} & Emolex & Emolex was built by crowdsourcing scenario. Emolex contains 14.182 words associated with eight primary emotion based on \cite{plutchik2001nature} : joy, sadness, anger, fear, trust, surprise, disgust, and anticipation.\\
Poria et. al. \cite{poria2013enhanced} & EmoSenticNet & EmoSenticNet(EmoSN) is an enriched version of SenticNet\cite{cambria2012senticnet}. Work by \cite{poria2013enhanced} added emotion label by mapping WordNet-Affect label to the SenticNet concepts. As a WordNet-Affect label, EmoSenticNet uses Six Ekman's basic emotion : anger, disgust, fear, happiness, sadness, and surprise. The whole list of this lexica contains 13.189 emotion labeled words.\\
Whissell et. al. \cite{whissell2009using} & Dictionary of Affect in Language (DAL) & DAL was developed by \cite{whissell2009using} and composed of 8742 English words. These words were labeled by three scores represent three emotion dimension : Pleasantness (the degree of pleasure), Activation (the degree of human response under some emotional condition), and Imagery (the degree of how easy the given words to be formed a mental picture).\\
Bradley et. al. \cite{bradley1999affective} & Affective Norms for English Words (ANEW) & This lexicon consists of 1,034 English words rated with ratings based on the Valence-Arousal-Dominance (VAD) model \cite{osgood_measurement_1957}.\\
de Albornoz et. al. \cite{de2012sentisense} & SentiSense & SentiSense  attaches emotional meanings to concepts from WordNet database. It is composed by a list of 5,496 words words labeled with emotional labels consisted of 14 emotional categories, which refer to a merge of Arnold, Plutchik and Parrot models.\\
Strapparava et. al. \cite{strapparava2004wordnet} & WordNet-Affect & WordNet-Affect is a lexical resource created based on WordNet which contains information about the emotions that the words convey. This lexicon is organized in six basic emotions: anger, disgust, fear, joy, sadness, surprise and, contains 2,874 synsets and 4,787 words\\
Saif M. Mohammad \cite{mohammad2018obtaining} & NRC VAD & The NRC VAD Lexicon is a list of English words and their valence, arousal, and dominance scores ranged between 0 (low) and 1 (high). This lexicon contains 20,000 terms and was built by crowdsourcing scenario.\\
Staiano and Guerini \cite{staiano2014depeche} & DepecheMood & DepecheMood is an emotion lexicon that crowdsource emotions through rappler website \footnote{\url{www.rappler.com}}. This lexicon contains 37,000 entries rated by  eight emotion categories (happy, sad, angry, afraid, annoyed, inspired, amused, and don't care).\\
Badaro et.al. \cite{badaro2018emowordnet} & EmoWordNet & EmoWordNet is created by expanding an existing emotion lexicon, DepecheMood \cite{staiano2014depeche}, by leveraging semantic knowledge from English WordNet. By alligning DepecheMood and WordNet, EmoWordNet consists of 67K terms, almost 1.8 times the size of DepecheMood. This lexicon has same emotion label as DepecheMood i.e., happy, sad, angry, afraid, annoyed, inspired, amused, and don't care.\\
\hline
\end{tabular}
\caption{\label{affective-resources-review} Summarization of the available affective resources for emotion classification task.}
\end{center}
\end{table*}

\section{Evaluating EAC}
We characterize the evaluation of Emotionally-Aware Chatbot into two different parts, qualitative and quantitative assessment. Qualitative assessment will focus on assessing the functionality of the software, while quantitative more focus on measure the chatbots' performance with a number.

\subsection{Qualitative Assessment}
Based on our investigation of several previous studies, we found that most of the works utilized ISO 9241 to assess chatbots' quality by focusing on the usability aspect. This aspect can be grouped into three focuses, including efficiency, effectiveness, and satisfaction, concerning systems' performance to achieve the specified goals. Here we will explain every focus based on several categories and quality attributes.

\subsubsection{Efficiency}
Efficiency aspect focuses on several categories, including robustness to manipulation and unexpected input \cite{thieltges2016devil}. Another study tries to asses the chatbots' ability to control damage and inappropriate utterance \cite{morrissey2013realness}.

\subsubsection{Effectiveness}
Effectiveness aspect covers two different categories, functionality and humanity. In functionality point of view, a study by \cite{eeuwen2017mobile} propose to asses how a chatbot can interpret command accurately and provide its status report. Other functionalities, such as chatbots' ability to execute the task as requested, the output linguistic accuracy, and ease of use suggested being assessed \cite{cohen2016oral}. Meanwhile, from the human aspect, most of the studies suggest each conversational machine should pass Turing test \cite{weizenbaum1966eliza}. Other prominent abilities that chatbot needs to be mastered can respond to specific questions and able to maintain themed discussion.

\subsubsection{Satisfaction}
Satisfaction aspect has three categories, including affect, ethics and behaviour, and accessibility. Affect is the most suitable assessment categories for EAC. This category asses several quality aspects such as, chatbots' ability to convey personality, give conversational cues, provide emotional information through tone, inflexion, and expressivity, entertain and/or enable the participant to enjoy the interaction and also read and respond to moods of human participant \cite{meira2015evaluation}. Ethic and behaviour category focuses on how a chatbot can protect and respect privacy \cite{eeuwen2017mobile}. Other quality aspects, including sensitivity to safety and social concerns and trustworthiness \cite{miner2016smartphone}. The last categories are accessibility, which the main quality aspect focus to assess the chatbot ability to detect meaning or intent and, also responds to social cues \footnote{\url{https://sloanreview.mit.edu/article/will-ai-create-as-many-jobs-as-it-eliminates/}}.

\subsection{Quantitative Assessment}
\subsubsection{Automatic Evaluation}
In automatic evaluation, some studies focus on evaluating the system at emotion level \cite{zhou2018emotional,li2019reinforcement}. Therefore, some common metrics such as precision, recall, and accuracy are used to measure system performance, compared to the gold label. This evaluation is similar to emotion classification tasks such as previous SemEval 2018 \cite{mohammad2018semeval} and SemEval 2019 \footnote{\url{https://www.humanizing-ai.com/emocontext.html}}. Other studies also proposed to use perplexity to evaluate the model at the content level (to determine whether the content is relevant and grammatical) \cite{rashkin2018know,lubis2018eliciting,li2019reinforcement}. This evaluation metric is widely used to evaluate dialogue-based systems which rely on probabilistic approach \cite{serban2016building}. Another work by \cite{rashkin2018know} used BLEU to evaluate the machine response and compare against the gold response (the actual response), although using BLEU to measure conversation generation task is not recommended by \cite{liu2016not} due to its low correlation with human judgment.

\subsubsection{Manual Evaluation}
This evaluation involves human judgement to measure the chatbots' performance, based on several criteria. \cite{zhou2018emotional} used three annotators to rate chatbots' response in two criteria, content (scale 0,1,2) and emotion (scale 0,1). Content is focused on measuring whether the response is natural acceptable and could plausible produced by a human. This metric measurement is already adopted and recommended by researchers and conversation challenging tasks, as proposed in \cite{shang2015neural}. Meanwhile, emotion is defined as whether the emotion expression contained in the response agrees with the given gold emotion category. Similarly, \cite{li2019reinforcement} used four annotators to score the response based on consistency, logic and emotion. Consistency measures the fluency and grammatical aspect of the response. Logic measures the degree whether the post and response logically match. Emotion measures the response, whether it contains the appropriate emotion. All of these aspects were measured by three scales 0, 1, and 2. Meanwhile, \cite{lubis2018eliciting} proposed naturalness and emotion impact as criteria to evaluate the chatbots' response. Naturalness evaluates whether the response is intelligible, logically follows the context of the conversation, and acceptable as a human response, while emotion impact measures whether the response elicits a positive emotional or triggers an emotionally-positive dialogue, since their study focus only on positive emotion. Another study by \cite{rashkin2018know} uses crowdsourcing to gather human judgement based on three aspects of performance including empathy/sympathy - did the responses show understanding of the feelings of the person talking about their experience?; relevance - did the responses seem appropriate to the conversation? Were they on-topic?; and fluency - could you understand the responses? Did the language seem accurate?. All of these aspects recorded with three different response, i.e., (1: not at all, 3: somewhat, 5: very much) from around 100 different annotators. After getting all of the human judgement with different criteria, some of these studies used a t-test to get the statistical significance \cite{li2019reinforcement,lubis2018eliciting}, while some other used inter-annotator agreement measurement such as Fleiss Kappa \cite{zhou2018emotional,rashkin2018know}. Based on these evaluations, they can compare their system performance with baseline or any other state of the art systems.

\section{Related Work}
There is some work which tried to provide a full story of chatbot development both in industries and academic environment. However, as far as my knowledge, there is still no study focused on summarizing the development of chatbot which taking into account the emotional aspect, that getting more attention in recent years. \cite{shawar2007chatbots} provides a long history of chatbots technology development. They also described several uses of chatbots' in some practical domains such as tools entertainment, tools to learn and practice language, information retrieval tools, and assistance for e-commerce of other business activities. Then, \cite{deshpande2017survey} reviewed the development of chatbots from rudimentary model to more advanced intelligent system. They summarized several techniques used to develop chatbots from early development until recent years. Recently, \cite{hussain2019survey} provide a more systematic shape to review some previous works on chatbots' development. They classify chatbots into two main categories based on goals, including task-oriented chatbots and non-task oriented chatbot. They also classify chatbot based on its development technique into three main categories, including rule-based, retrieval-based, and generative-based approach. Furthermore, they also summarized the detailed technique on these three main approaches.

\section{Discussion and Conclusion}
In this work, a systematic review of emotionally-aware chatbots is proposed. We focus on three main issues, including, how to incorporate affective information into chatbots, what are resources that available and can be used to build EAC, and how to evaluate EAC performance. The rise of EAC was started by Parry, which uses a simple rule-based approach. Now, most of EAC are built by using a neural-based approach, by exploiting emotion classifier to detect emotion contained in the text. In the modern era, the development of EAC gains more attention since Emotion Generation Challenge shared task on NLPCC 2017. In this era, most EAC is developed by adopting encoder-decoder architecture with sequence-to-sequence learning. Some variant of the recurrent neural network is used in the learning process, including long-short-term memory (LSTM) and gated recurrent unit (GRU). There are also some datasets available for developing EAC now. However, the datasets are only available in English and Chinese. These datasets are gathered from various sources, including social media, online website and manual construction by crowdsourcing. Overall, the difference between these datasets and the common datasets for building chatbot is the presence of an emotion label. In addition, we also investigate the available affective resources which usually use in the emotion classification task. In this part, we only focus on English resources and found several resources from the old one such as LIWC and Emolex to the new one, including DepecheMood and EmoWordNet. In the final part, we gather information about how to evaluate the performance of EAC, and we can classify the approach into two techniques, including qualitative and quantitative assessment. For qualitative assessment, most studies used ISO 9241, which covers several aspects such as efficiency, effectiveness, and satisfaction. While in quantitative analysis, two techniques can be used, including automatic evaluation (by using perplexity) and manual evaluation (involving human judgement). Overall, we can see that effort to humanize chatbots by incorporation affective aspect is becoming the hot topic now. We also predict that this development will continue by going into multilingual perspective since up to now every chatbot only focusing on one language. Also, we think that in the future the studies of humanizing chatbot are not only utilized emotion information but will also focus on a contextual-aware chatbot.


\end{document}